\documentclass[conference]{IEEEtran}
\IEEEoverridecommandlockouts
\usepackage{cite}
\usepackage{amsmath,amssymb,amsfonts}
\usepackage{algorithm}
\usepackage{graphicx}
\usepackage{algpseudocode}
\usepackage{subcaption}
\usepackage{textcomp}
\usepackage{xcolor}
\def\BibTeX{{\rm B\kern-.05em{\sc i\kern-.025em b}\kern-.08em
    T\kern-.1667em\lower.7ex\hbox{E}\kern-.125emX}}
    
\usepackage{makecell}

\usepackage{booktabs}

\usepackage[caption=false]{subfig}

\newcolumntype{P}[1]{>{\centering\arraybackslash}p{#1}}
\newcolumntype{M}[1]{>{\centering\arraybackslash}m{#1}}
\begin{document}

\title{Benchmarking Spiking Neural Network Learning Methods with Varying Locality}

\author{Jiaqi Lin, Sen Lu, Malyaban Bal, Abhronil Sengupta
\thanks{J. Lin, S. Lu, M. Bal and A. Sengupta are with the School of Electrical Engineering and Computer Science, The Pennsylvania State University, University Park, PA 16802, USA. E-mail: jkl6467@psu.edu, sengupta@psu.edu.}}

\maketitle

\begin{abstract}
Spiking Neural Networks (SNNs), providing more realistic neuronal dynamics, have been shown to achieve performance comparable to Artificial Neural Networks (ANNs) in several machine learning tasks. Information is processed as spikes within SNNs in an event-based mechanism that significantly reduces energy consumption. However, training SNNs is challenging due to the non-differentiable nature of the spiking mechanism. Traditional approaches, such as Backpropagation Through Time (BPTT), have shown effectiveness but come with additional computational and memory costs and are biologically implausible. In contrast, recent works propose alternative learning methods with varying degrees of locality, demonstrating success in classification tasks. In this work, we show that these methods share similarities during the training process, while they present a trade-off between biological plausibility and performance. Further, given the implicitly recurrent nature of SNNs, this research investigates the influence of the addition of explicit recurrence to SNNs. We experimentally prove that the addition of explicit recurrent weights enhances the robustness of SNNs. We also investigate the performance of local learning methods under gradient and non-gradient-based adversarial attacks.
\end{abstract}

\section{Introduction}
Spiking Neural Networks (SNNs) are a type of Artificial Neural Networks (ANNs) that are inspired by biological neurons in the brain \cite{maass1997networks, gerstner2002spiking}. In contrast to traditional ANNs, SNNs use a more biologically realistic model of neuron behavior, where information is processed through the propagation of spikes between neurons in the network \cite{tavanaei2019deep}. The timing and rate of spikes are influenced by the strength of synaptic connections between neurons, which can be modified through a process called synaptic plasticity \cite{diehl2015unsupervised, bi1998synaptic}. In SNNs, neuronal activities are based on discrete spikes, which occur only when the membrane potential of a neuron exceeds a certain threshold. This event-driven processing operates at significantly lower energy levels than traditional computing architectures \cite{mead1990neuromorphic, merolla2014million, benjamin2014neurogrid}, as the neuronal and synaptic state updates consume power only when they are actively spiking \cite{boahen2006neuromorphic}. Neuromorphic hardware implementations, like Intel's Loihi \cite{davies2018loihi} and the SpiNNaker project \cite{furber2014spinnaker}, are specifically designed to take advantage of SNNs. The energy efficiency offered by SNNs integrated with neuromorphic hardware provides efficient possibilities for deploying machine learning applications in power-constrained environments, such as mobile devices, wearable technology, and remote sensing systems, among others. 

Additionally, since membrane potentials are accumulated over time retaining information from previous time steps, SNNs are inherently recurrent \cite{ponghiran2022spiking}. This characteristic of SNNs allows them to naturally process time-dependent data, rendering them suitable for tasks such as speech recognition or time-series prediction \cite{ponghiran2022spiking, lsnn_bellec2018long, yin2021accurate, eprop_bellec2020solution,bal2023spikingbert}. SNNs with explicit recurrent connections have been shown to further augment their performance in speech processing tasks \cite{lsnn_bellec2018long}. 

Although SNNs are demonstrated to provide multiple benefits from algorithms to hardware by leveraging bio-realistic computations, Backpropagation Through Time (BPTT), which is widely used to train SNNs, is not biologically plausible. 
Firstly, BPTT assumes weight symmetry for the forward and backward passes, which is undesirable in biological neural networks \cite{lillicrap2016random}. Secondly, BPTT requires global error propagation, which is inconsistent with the local learning mechanisms observed in biological neural networks \cite{ baldi2017learning, decolle_kaiser2020synaptic}. 
Moreover, the explicit calculation and back-propagation of gradients and error signals in BPTT are not directly observed in the brain \cite{crick1989recent, lillicrap2020backpropagation}. 
Lastly, BPTT maintains a history of the membrane potential and spike history of each neuron over time. This process can be memory-intensive and computationally expensive \cite{werbos1990backpropagation, williams1995gradient, decolle_kaiser2020synaptic, eprop_bellec2020solution,lu2022neuroevolution}. 
To solve this problem, local learning alternatives to BPTT have been proposed for training SNNs \cite{lsnn_bellec2018long, zenke2018superspike, neftci2019surrogate, decolle_kaiser2020synaptic,bal2022sequence}.
These methods aim to address the challenges associated with training SNNs while still maintaining biologically plausible learning rules and enabling more efficient and accurate learning in spiking networks.

\begin{figure*}[htbp] \centerline{\includegraphics[width=0.8\textwidth]{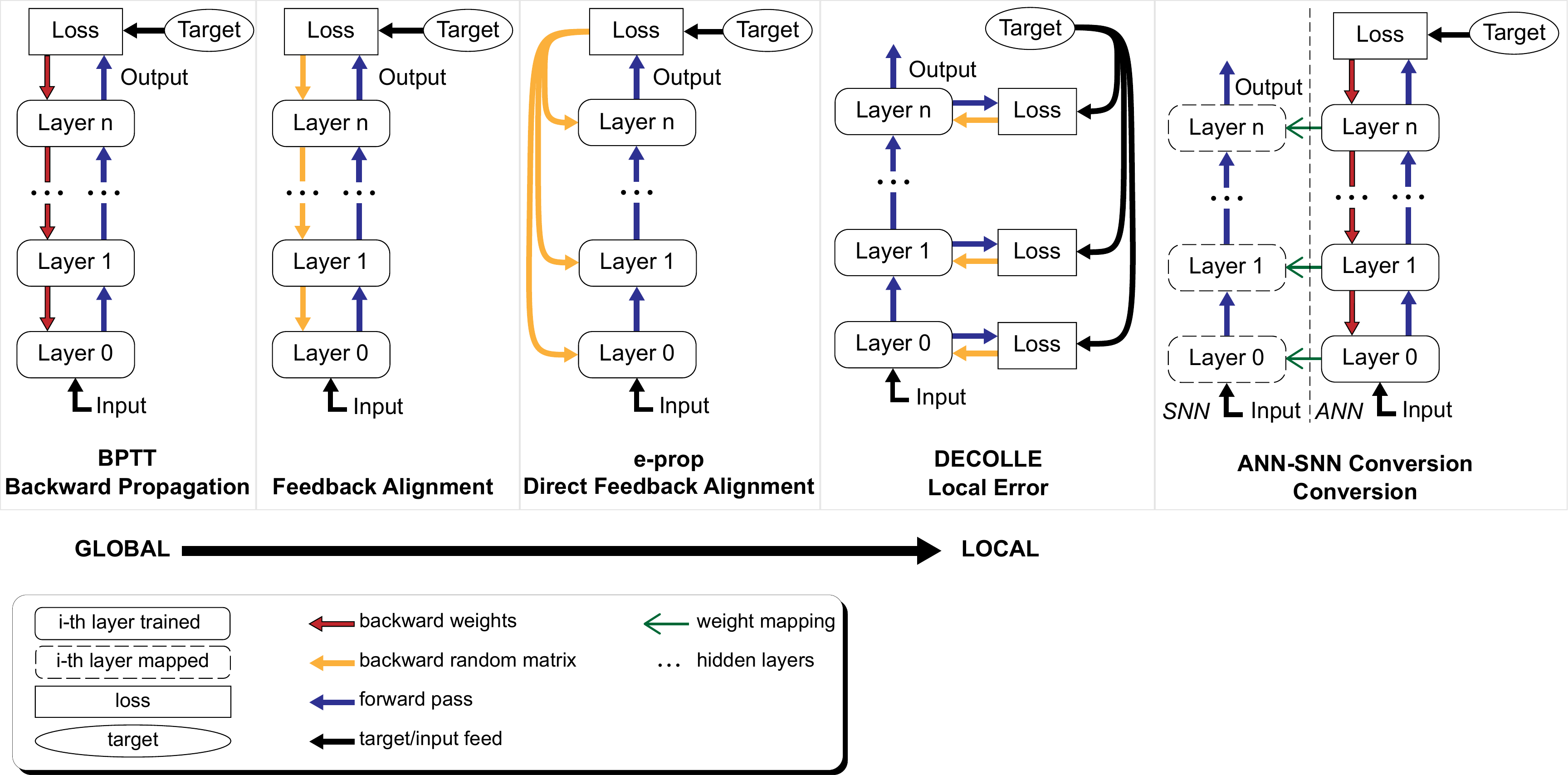}} \caption{SNN training methods considered in this study from global learning to local learning: BPTT, Feedback Alignment, e-prop, DECOLLE. ANN-SNN conversion is also depicted on the extreme right.} \label{fig:lr_mtd} \end{figure*}

Based on the locality of training methods \cite{neftci2019surrogate}, we revisit these alternatives to BPTT. Figure \ref{fig:lr_mtd} distinguishes SNN training methods based on varying levels of locality. Beginning with BPTT, which is a global learning method, it unrolls neural dynamics over time, and it involves backpropagating errors and learning signals through layers. The connection used for backpropagation is the same weight used for forward propagation. An alternative, named feedback alignment, replaces the symmetric weight connections with random matrices during backward passes. This method removes the need for symmetric weights during learning. E-prop introduces a type of direct feedback alignment that provides higher biological plausibility  \cite{eprop_bellec2020solution}. In this method, the error calculated at the output layer is propagated directly to each layer without propagating errors layer by layer. To obtain better performance, e-prop employs symmetric weights in the process of direct propagation. We next consider a more localized method, DECOLLE, in the regime of local error propagation \cite{decolle_kaiser2020synaptic}. In this learning rule, errors are propagated at each layer. A cost function is assigned at each layer by a random matrix mapping the output from each layer to the pseudo-targets. In general, all these learning rules provide bio-plausible alternatives to global learning rules. 


\section{Related Works}

\subsection{Performance Evaluation}

Recently, the development of novel learning algorithms to train SNNs has gained traction, propelled by  SNNs' unique spiking nature and energy advantage over ANNs. Whether energy efficiency is a compromise of performance remains a question. Ref. \cite{deng2020rethinking} explored such a co-design evaluation with respect to ANN performance. 
Their results indicated that SNNs maintained acceptable accuracy drops on simpler visual tasks like MNIST classification with much better energy efficiency, while in relatively more complicated visual tasks like CIFAR10 classification, the accuracy of SNNs degraded further. Despite this, SNNs evaluated on neuromorphic datasets were illustrated to outperform ANNs.
Subsequent progress in SNN training has produced state-of-the-art performance of SNNs in several image recognition benchmarks, with recent works indicating highly similar representations learned by residual SNNs and ANNs using representational similarity measurements \cite{ckasnn_li2023uncovering}. 

Moreover, the inherently recurrent nature of SNNs resembles the operation of recurrent neural networks (RNNs) in terms of temporal and spatial neural dynamics. Inspired by such characteristics, prior work has compared SNNs and RNNs, highlighting the superior performance of SNNs over RNNs on neuromorphic datasets \cite{he2020comparing}. Additionally, the work investigated the influence of temporal resolution, leakage, and reset mechanisms on the performance of SNNs, illustrating that the exclusion of leakage and reset mechanisms negatively impacted the performance, and the increment of temporal resolution was not necessarily useful. 
In these benchmarking works, SNNs were trained with the BPTT training method, 
 where surrogate functions were used to overcome the non-differentiability of spiking neurons.
To quantitatively analyze the impact of surrogate functions on the performance of SNNs, experimental results from previous work have demonstrated that surrogate functions help recurrent SNNs outperform conventional RNNs, and the choice of surrogate functions exhibits negligible impact on performance \cite{yin2021accurate}. Additionally, this work examined the advantage of adaptive neurons \cite{lsnn_bellec2018long}, where thresholds are updated based on firing rates, and illustrated that recurrent SNNs with adaptive neurons gain performance advantages over SNNs with fixed thresholds. Mixed usage of both types of neurons was shown to improve the performance of SNNs over LSTM architectures \cite{lsnn_bellec2018long}.

In conclusion, prior works have mainly targeted benchmarking comparable performance of SNNs and ANNs. Some works have investigated the influence of various SNN-specific control knobs like leakage, temporal dynamics, among others, on SNN performance. However, the majority of benchmarking efforts optimize SNNs via the BPTT algorithm, thereby not providing insights related to local learning methods. In contrast to prior works, we perform an extensive benchmarking analysis with a complementary perspective where we investigate the performance of SNNs with regard to local learning methods in terms of accuracy and robustness with both implicit and explicit recurrent architectures.

\subsection{Adversarial Attacks}

The adversarial robustness of SNNs has been proven by various researchers. Sharmin et al. \cite{sharmin2019comprehensive} explored gradient-based attacks and their variants on both the ANN-converted SNN and SNN trained from scratch and found SNN's robustness against adversarial attacks. Similarly, Liang et al. \cite{liang2021exploring} conducted untargeted gradient attacks and found that SNNs require more perturbations for successful attacks. Marchisio et al. \cite{marchisio2020spiking} considered a deep Spiking Deep Belief Network (SDBN) using noise attacks with the same motive of backdoor trigger attacks \cite{bd_badnet_gu2019badnets} and reached similar conclusions.
However, their studies were either focused on a single training method or relied entirely on the gradient. While at its exploration stage, many recent SNN training methods look for alternatives to using local training methods \cite{lu2022neuroevolution,bhargava2022gradient}. In this work, we aim to bridge this gap by revisiting adversarial attacks on SNNs trained by local learning methods and provide a benchmarking analysis for non-gradient poisoning attacks as well. Although adversarial training strategies strengthen machine learning models against adversarial inputs in the domain of SNNs \cite{kundu2021hire, ding2022snn, ozdenizci2023adversarially}, these mechanisms are out of the scope of this study. 

The primary contributions of this work are the following: 
\begin{itemize}
    \item We extensively benchmark SNNs' performance, concentrating on learning methods such as BPTT, e-prop, and DECOLLE applied to both implicit and explicit recurrent architectures. We highlight the trade-off between biological plausibility and performance, as well as the enhancement of accuracy achieved by incorporating additional weights in explicit recurrent SNNs.
    \item We perform an in-depth analysis to reveal that training methods with similar locality share representational similarities. Additionally, we find recurrent and linear weights have comparable importance, with each linear layer's significance growing from input to output.
    \item This study offers insights into the robustness of local learning methods by revisiting adversarial attacks for both gradient-based and non-gradient poisoning attacks. SNNs trained with local learning methods are more robust than those using global learning methods for gradient-based attacks. Analysis shows that explicit recurrent SNNs have greater robustness, supported by the Centered Kernel Alignment (CKA) metric. 
    
\end{itemize}

\section{Methods}
Sections A-B discuss SNN computational models and learning algorithms considered in this work. {Section C analyzes the computational complexity of learning methods.} Sections D-E introduce empirical tools used to understand the learning dynamics of various SNN topologies and learning methods.

\subsection{Spiking Neurons}
In this study, we use SNNs with Leaky-Integrate-and-Fire (LIF) neurons \cite{yin2021accurate, neftci2019surrogate}. The membrane potential $u^{l,t}$ and synaptic currents $\iota^{l,t}$ for a layer $l$ at time step $t$ in a discrete time setting are expressed as:
\begin{equation}
\label{eq:n}
\begin{split}
\iota^{l,t} &= \exp(-\frac{1}{\tau_{\rm syn}}) \iota^{l,t-1} + w^l s^{l-1,t} + v^l s^{l,t-1} \\
u^{l,t} &= \exp(-\frac{1}{\tau_{\rm mem}}) u^{l,t-1} + \iota^{l,t} - R\mathcal{V}_{\rm th}
\end{split}
\end{equation}
Here, $w^l$ represents the weight connection between the current layer $l$ and the previous layer $l-1$, and $v^l$ represents the recurrent weight connection of the current layer. $s^{l-1,t}$ denotes the activation of the previous layer $l-1$ at time step $t$. 
The parameters $\exp(\frac{-1}{\tau_{\rm syn}})$ and $\exp(\frac{-1}{\tau_{\rm mem}})$ correspond to the synaptic and membrane decay rates respectively. 
Each neuron maintains a membrane potential $u^{l,t}$, and if it surpasses the threshold $\mathcal{V}_{\rm th}$ of the spiking neuron, a spike will be fired. 
The membrane potential $u^{l,t}$ is subtracted by the threshold $\mathcal{V}_{\rm th}$ after each firing event, which is captured by the refractory period $R$.

\subsection{Training Methods}
Training a neural network consists of steps to optimize the loss function $L$ with respect to the parameter set $w$. In this section, we will introduce typical training algorithms to train an SNN. 

\subsubsection{Backpropagation Through Time (BPTT)}

Backpropagation Through Time (BPTT) is an optimization algorithm specifically designed for training Recurrent Neural Networks (RNNs) as an extension of the standard backpropagation algorithm to handle time-dependent sequences and recurrent connections \cite{werbos1990backpropagation,rumelhart1986learning}. As SNNs are inherently recurrent in time, training SNNs through BPTT is similar to training RNNs with BPTT. The algorithm functions by unfolding the network through time to capture the temporal dependencies in the input data and applying the backpropagation algorithm over time. The weight update formula for BPTT can be written as follows:
\begin{equation}
\begin{split}
    \Delta w_{ij} &= -\eta \sum_{t=1}^T \frac{\partial L}{\partial s_{ij}^t} \frac{\partial  s_{ij}^t}{\partial w_{ij}} = -\eta \sum_{t=1}^T  \frac{\partial L}{\partial s_i^{t}} \frac{\partial s_i^{t}}{\partial u_i^{t}} \frac{\partial u_i^{t}}{\partial w_{ij}} \\
    &= -\eta \sum_{t=1}^T  \frac{\partial L}{\partial s_i^{t}} \sigma^\prime(u_i^{t})  \frac{\partial u_i^{t}}{\partial w_{ij}} 
\end{split}
\end{equation}
where, $T$ is the total number of time steps, $\eta$ is the learning rate, and $L$ is the loss function. At time step $t$, $w_{ij}$ represents the weight connections between neurons $i$ and $j$, $s_i^{t}$ is the activation of neuron $i$, and $\sigma^\prime(u_i^{t})$ approximates the derivative of the activation function with respect to the pre-activation value $u_i^{t}$ for neuron $i$ with a surrogate function.

\subsubsection{E-prop}
Recent studies of error propagation methods in SNNs propose a new method called e-prop \cite{eprop_bellec2020solution}. Compared with BPTT, e-prop offers higher biological plausibility, as it is inspired by the concept of synaptic eligibility traces observed in neuroscience \cite{EligibilityTraces}, which weigh the temporal differences of presynaptic and postsynaptic neuronal activations. While BPTT relies on a global learning rule that requires error information to be propagated backward through the network between layers, e-prop aims to derive a local learning rule, in which the error signals are propagated directly from the output layer to hidden units. The local learning rule in e-prop makes it more suitable for neuromorphic hardware implementations.

In the e-prop method, the weight updates are based on the product of eligibility traces and learning signals. The eligibility trace $e_{ij}^t$ for a synapse connecting neuron $i$ to neuron $j$ at time step $t$ can be computed recursively by eligibility vector $\upsilon$:
\begin{equation}
    \begin{split}
        \label{eq:et1}
e_{ij}^t &= \frac{\partial s_i^t}{\partial w_{ij}} = \frac{\partial s^t_i}{\partial u^t_i}\upsilon_{ij}^t \\
\upsilon^t_{ij} &= \frac{\partial u^t_i}{\partial u^{t-1}_i}\upsilon^{t-1}_{ij} + \frac{\partial u^t_i}{\partial w_{ij}}
    \end{split}
\end{equation}
The eligibility trace captures the recent history of a synapse's contribution to the post-synaptic neuron's activation. The learning signal $L_i^t$ for neuron $i$ at time step $t$, following direct feedback alignment, routes the error calculated at the output layer to the current neuron:
\begin{equation}
\label{eq:el}
L_i^t = \sum\limits_k g_{ik} \frac{\partial L_k}{\partial s^t_k}
\end{equation}
where $g_{ik}$ is a fixed random matrix and $\partial L_k/\partial s^t_k$ captures the loss calculated at the output layer. The weight update can be expressed as:
\begin{equation}
\label{eq:ewu}
\Delta w_{ij} = -\eta \sum_{t=1}^{T} L_i^t \cdot e_{ij}^t
\end{equation}

\subsubsection{DECOLLE}
Deep Continuous Local Learning (DECOLLE) is a biologically plausible online learning algorithm for training SNNs \cite{decolle_kaiser2020synaptic}. DECOLLE combines the principles of local learning with deep architectures, enabling the training of SNNs in a layer-by-layer manner without the need for backpropagation or weight transport.
In DECOLLE, each layer in the network learns features using local information, which allows the network to adapt online without relying on global error signals. This is achieved by attaching an auxiliary cost function to the random readouts at each layer. The layer-specific random readouts $y^t_i$ at time $t$ are calculated as:
\begin{equation}
\label{eq:dl}
    \begin{split}
    \frac{\partial L^t_i}{\partial s_i^{t}} &= \frac{\partial L(y^t_i, \hat{y}_i)}{\partial s_i^{t}} \\
    y_i^t &= g_i s_i^t 
\end{split}
\end{equation}
where $g_i$ is a fixed random matrix for layer $l$. Then, the loss $L$ is calculated as the sum of layerwise differences between readouts $y^t_i$ and pseudo-target $\hat{y}_i$. Based on this, weight updates at each layer for a specific time step $t$ are determined using a local learning rule:
\begin{equation}
    \begin{split}
    \Delta w_{ij}^t &= -\eta \frac{\partial L^t_i}{\partial w_{ij}^t} = -\eta \frac{\partial L^t_i}{\partial s_i^t}\frac{\partial s_i^t}{\partial w_{ij}^t} \\
    \frac{\partial s_i^t}{\partial w_{ij}^t} &= \sigma^\prime(u_i^t) (p_i^t - \rho\frac{\partial r_i^t}{\partial w_{ij}^t})
\end{split}
\end{equation}
where $p_i^t$ refers to the traces of the membrane potential of neuron $i$ driven solely by incoming spikes, $r_i^t$ with a constant factor $\rho$ that captures the refractory dynamics depending on the spiking history of the neuron $i$. DECOLLE ignores the spiking history dependencies, considering $r_i^t$ to have a negligible impact on the membrane potential. Consequently, the synaptic weight updates become:
\begin{equation}
\label{eq:dwu}
\Delta w_{ij} =-\eta \sum_{t=1}^{T}\frac{\partial L^t_i}{\partial s_i^t}\sigma^\prime(u_i^t)p_i^t
\end{equation}
Ignoring refractory dynamics ensures that the learning process relies exclusively on local information, adhering to the principles of biologically plausible learning rules.

 \subsection{Computational Advantages of Local Learning Methods \label{sec:enr}}
\begin{table}[htbp]
\caption{Complexity analysis of gradient computation for different training methods at a specific layer \cite{decolle_kaiser2020synaptic}. Here, we assume neurons between layers are fully connected. $N_{\rm in}$ is the input size, $N_{\rm neu}$ represents the number of neurons in current layer, $T$ is time steps, $N_{\rm ro}$ indicates the number of readout neurons in DECOLLE. The calculations below does not account for the gradient accumulation overhead in epoch-wise learning.\label{tab:comp}}
\begin{center}
\begin{tabular}{|l|l|l|}
\hline
\textbf{Method}& Space & Time \\
\hline
    \hline
    BPTT &  $O(N_{\rm in}T)$ & $O(N_{\rm in}N_{\rm neu}T)$\\
     e-prop & $O(N_{\rm in}N_{\rm neu})$ & $O(N_{\rm in}N_{\rm neu})$ \\
     DECOLLE &  $O(1)$ &  $O(N_{\rm neu}N_{\rm ro} +N_{\rm in}N_{\rm neu})$ \\
    \hline
\end{tabular}
\end{center}
\end{table}

Table \ref{tab:comp} provides the computational complexity for the aforementioned training methods. BPTT performs backpropagation on unrolled $N_{\rm in}$ neuron states across time $T$, resulting in a space complexity of $O(N_{\rm in}T)$ and a time complexity of $O(N_{\rm in}N_{\rm neu}T)$, as it multiplies error signals with gradients computed in the current layer with $N_{\rm neu}$ neurons over $T$.
The e-prop learning rule maintains $N_{\rm in}N_{\rm neu}$ number of eligibility traces (Equation \ref{eq:et1}) in forward manner for each synaptic connection, leading to a space complexity of $O(N_{\rm in}N_{\rm neu})$ \cite{eprop_bellec2020solution}. During the gradient computation, the error signals of size $N_{\rm neu}$ are multiplied by the eligibility traces (Equation \ref{eq:ewu}) with a time complexity of $O(N_{\rm in}N_{\rm neu})$.
DECOLLE performs online weight updates at each time step \cite{decolle_kaiser2020synaptic}, using available neuron states without storing, resulting in a space complexity of $O(1)$. 
During the weight gradient computation, local errors of size $N_{\rm neu}$ are computed (Equation \ref{eq:dl}) using $N_{\rm ro}N_{\rm neu}$ multiplication operations for $N_{\rm neu}$ neurons. Then these local errors are multiplied by the number of inputs $N_{\rm in}$, yielding a total time complexity of $O(N_{\rm neu}N_{\rm ro} +N_{\rm in}N_{\rm neu})$.

\subsection{Fisher information}
In statistical modeling, Fisher information measures the quantity of information that a given data sample holds about an unknown parameter that the data sample depends on \cite{fisher1925theory}. Previous work has applied Fisher information to analyze the learning dynamics of SNNs \cite{fisher_kim2023exploring}. 

As SNNs rely on accumulating input data over multiple time steps to predict class probabilities, the amount of Fisher information Matrix (FIM) $M_t$ accumulated in SNNs at a given time step is the sum of information across all previous time steps, from 1 to $t$.
\begin{equation}
    \begin{split}
M_t&=\mathbb{E}_{x\sim D}\mathbb{E}_{y\sim f_w(y|x_{i\leq t})} \\ 
&\left[\nabla_w \log f_w (y|x_{i\leq t})\nabla_w \log f_w(y|x_{i\leq t})^\intercal\right] 
\end{split} 
\end{equation}
Here, $x$ is input image sampled from data distribution $D$, $y$ is the output variable, and $i \in \{1, \ldots, t\}$ represents the index of the time step. Direct calculation on the Fisher information Matrix (FIM) is computationally expensive, due to the large number of parameters in an SNN. Consequently, computing the trace of the FIM is a more efficient alternative \cite{fisher_kim2023exploring}. Given a dataset of $N$ training samples, we can compute the Fisher information at time $t$ as follows:
\begin{equation}
F_t = \frac{1}{N} \sum\limits_{n = 1}^N \lVert\nabla_w 
\log f_w(y|x^n_{i\leq t}) \rVert^2
\end{equation}
We will use this metric to quantify the relative importance of recurrent weights in later sections.

\subsection{Centered Kernel Alignment (CKA)}
CKA is used to quantify the similarity between representations in two arbitrary layers through the normalized Hilbert-Schmidt Independence Criterion (HSIC) \cite{ckann_kornblith2019similarity, ckasnn_li2023uncovering}. In this work, the CKA calculation is adopted from Li et al. \cite{ckasnn_li2023uncovering}.

Let $\mathbf{\Gamma}_\alpha \in \mathbb{R}^{b\times T p_\alpha}$ and $\mathbf{\Gamma}_\beta \in \mathbb{R}^{b\times T p_\beta}$ be the representations of layer $\alpha$ of one SNN with $p_\alpha$ hidden neurons and layer $\beta$ of another SNN with $p_\beta$ hidden neurons over $T$ time steps, where $b$ is the batch size. The representations across all time steps are concatenated in SNNs. Then CKA is defined as:
\begin{equation}
CKA(\mathbf{K}, \mathbf{L}) = \frac{HSIC(\mathbf{K},\mathbf{L})}{\sqrt{HSIC(\mathbf{K},\mathbf{K})HSIC(\mathbf{L},\mathbf{L})}}.
\end{equation}
where, $\mathbf{K} = \mathbf{\Gamma}_\alpha\mathbf{\Gamma}_\alpha^\intercal$ and $\mathbf{L} = \mathbf{\Gamma}_\beta\mathbf{\Gamma}_\beta^\intercal$ are the Gram matrices with shape $b\times b$ representing the similarity between examples. The HSIC is a statistical measure that assesses the independence of two sets of variables \cite{greenfeld2020robust}. HSIC is defined as following: 
\begin{equation}
\begin{split}
  HSIC(\mathbf{K},\mathbf{L}) &= \frac{1}{(b-1)^2}tr(\mathbf{KCLC}) \\  
  \mathbf{C} &= \mathbf{J} - \frac{1}{b} \mathbf{O}
\end{split}
\end{equation}
where, $\mathbf{J}$ is the identity matrix and $\mathbf{O}$ is a matrix consisting of all 1's. HSIC score of 0 means independence between two variables. CKA normalizes this into a similarity index from 0 to 1, where higher values imply greater similarity. We adhere to the use of an unbiased estimator for calculating HSIC across mini-batches \cite{ckasnn_li2023uncovering, nguyen2020wide, song2012feature}. CKA metric will also be used in later sections to empirically explain the performance differences of various SNN learning methods in terms of accuracy and robustness.

\section{Results}

\subsection{Experimental Setup}
Experiments conducted in this study include Neuromorphic-MNIST (N-MNIST), DVS Gesture, and TIMIT for the investigation of the performance of various training methods.

\textbf{N-MNIST:} The N-MNIST dataset is a neuromorphic adaptation of the handwritten digits classification task \cite{orchard2015NMNIST}. In contrast to the MNIST dataset \cite{deng2012mnist}, which consists of static images, N-MNIST encapsulates changes in pixel values as events unfold over time. N-MNIST is specifically designed for the evaluation of neuromorphic algorithms and models \cite{lee2016training, decolle_kaiser2020synaptic}. 

\textbf{DVS Gesture:} The DVS Gesture dataset \cite{amir2017low} contains data from 29 subjects under three lighting conditions. A time step of 60 is used to evaluate performance \cite{he2020comparing}.

\textbf{TIMIT:} The TIMIT dataset is a speech and audio processing sequence-to-sequence task \cite{garofolo1993darpa, lsnn_bellec2018long, eprop_bellec2020solution}. Although N-MNIST, DVS Gesture, and TIMIT datasets all involve the temporal dimension, data in TIMIT exhibit a higher degree of dependency between time steps.

It is worth mentioning here that the e-prop method is currently limited to linear layers \cite{eprop_bellec2020solution}. To ensure an equitable comparison between training methods without performance degradation, we therefore restrict our main experiments to fully connected architectures. To assess the generalizability of our conclusions, we further evaluate convolutional architectures trained with BPTT and DECOLLE in Appendix \ref{app:perf}.
In this work, performance and robustness measurements are conducted for two types of SNN architectural designs. One is a feed-forward SNN (FF) and another is a recurrent SNN (REC) architecture, which introduces additional weights in the linear layers in FF. To classify the N-MNIST dataset, both FF and REC SNNs consisting of two hidden layers of 120 and 84 neurons are used. The DVS Gesture dataset is classified by FF and REC SNNs with one hidden layer of 512 neurons. The TIMIT classification task utilizes FF and REC SNNs with one hidden layer of 400 neurons. Notably, due to the configuration of e-prop, the output layer does not contain an explicit recurrent weight connection.

To ensure a fair comparison between the training methods, the hyperparameters of the SNN models trained with different paradigms were optimized for the best performance (Appendix \ref{app:param}). The experiments were implemented using the snnTorch and PyTorch libraries in Python and conducted on an Nvidia RTX 2080 Ti GPU with 11GB of memory.

\subsection{Performance Evaluation}
\label{sec:perf}
\begin{table}[htbp]
\caption{SNN accuracy (\%) reported in prior works on the N-MNIST, DVS Gesture, and TIMIT datasets. Diverse architectures and datasets hinder direct comparison of each training method.}
\begin{center}
\begin{tabular}{|c|c|c|c|c|c|c|}
\hline
\textbf{Training}&\multicolumn{2}{|c|}{\textbf{N-MNIST}} &\multicolumn{2}{|c|}{\textbf{DVS Gesture}} &\multicolumn{2}{|c|}{\textbf{TIMIT}}\\
\cline{2-7} 
\textbf{Methods} & \textit{FF} & \textit{REC} & \textit{FF} & \textit{REC} & \textit{FF} & \textit{REC} \\
\hline
 BPTT    & $98.3^{\mathrm{a}}$ && $87.5^{\mathrm{a}}$&  & $57.7^{\mathrm{b}}$& $67.1^{\mathrm{c}}$  \\ 
 \hline
e-prop   &  & & & & $55.2^{\mathrm{d}}$ & $65.4^{\mathrm{e}}$  \\ 
\hline
DECOLLE & $99.0^{\mathrm{f}} $ &    & $95.5^{\mathrm{f}} $&  &  &  \\ 
\hline
\multicolumn{7}{l}{\parbox [t] {220pt}{$^{\mathrm{a}}$This architecture uses one hidden layer with 512 neurons\cite{he2020comparing}.}} \\
\multicolumn{7}{l}{\parbox [t] {220pt}{$^{\mathrm{b}}$This architecture uses 3 hidden layers each with 900 neurons in an LSNN architecture \cite{lsnn_bellec2018long}, which is an LSTM-equivalent architecture in the spiking domain. In general, LSNN has better performance compared to recurrent SNN. The performance is obtained from \cite{eprop_bellec2020solution}.}} \\
\multicolumn{7}{l}{\parbox [t] {220pt}{$^{\mathrm{c}}$This architecture uses one hidden layer with 400 neurons using an LSNN architecture \cite{eprop_bellec2020solution}.}} \\

\multicolumn{7}{l}{\parbox [t] {220pt}{$^{\mathrm{d}}$With 1 hidden layer of 1200 neurons, a feed-forward LSNN trained with e-prop achieves $49.94\%$ accuracy; With 3 hidden layers, each of which consists of 900 neurons, a feed-forward LSNN trained with e-prop achieves $55.2\%$ accuracy \cite{eprop_bellec2020solution}.}} \\
\multicolumn{7}{l}{\parbox [t] {220pt}{$^{\mathrm{e}}$This architecture uses one hidden layer with 400 neurons using an LSNN architecture \cite{eprop_bellec2020solution}.}} \\
\multicolumn{7}{l}{\parbox [t] {220pt}{$^{\mathrm{f}}$This architecture consists of 3 convolutional layers with 64, 128, and 128 channels respectively \cite{decolle_kaiser2020synaptic}. }}
\end{tabular}
\label{tab:prev_perf}
\end{center}
\end{table}

Table \ref{tab:prev_perf} presents the outcomes of previous studies that have benchmarked the performance of SNNs. These results show promising performance of SNNs in several classification tasks in comparison to state-of-the-art ANNs. However, due to the lack of consistency in neural network architecture designs and variations in training sets, direct comparison between different training methods is infeasible. To address this problem, we have conducted a series of benchmark tests with consistent neural network architectures and training datasets. This work aims to offer a more comprehensive and insightful assessment of the performance of local learning methods and the impact of explicit recurrence on SNNs.

\begin{table}[htbp]
\caption{SNN accuracy (\%) is benchmarked on the N-MNIST, DVS Gesture, and TIMIT datasets, with results averaged across five independent runs. Performance degradation in local learning methods highlights a trade-off between biological plausibility and performance.}
\begin{center} 
\begin{tabular}{|c|c|c|c|c|c|c|c|}
\hline
\textbf{Training}&\multicolumn{2}{|c|}{\textbf{N-MNIST}} &\multicolumn{2}{|c|}{\textbf{DVS Gesture}} &\multicolumn{2}{|c|}{\textbf{TIMIT}}\\
\cline{2-7} 
\textbf{Methods} & \textit{FF} & \textit{REC} & \textit{FF} & \textit{REC} & \textit{FF} & \textit{REC} \\
\hline
BPTT    & 97.12  & 97.77  & 87.73 & 89.23 & 52.31 & 56.93\\
\hline
e-prop   & 96.44 & 96.80 & 86.15 & 87.20 & 51.12 & 54.99 \\
\hline
DECOLLE & 93.61 & 90.52 & 84.16 & 85.17 &  42.30 & 45.70 \\ 
\hline
\end{tabular}
\label{tab:bench}
\end{center}
\end{table}

In our investigations, we have observed that the introduction of recurrent weights in the REC architecture yields significant performance improvements when classifying the DVS Gesture and TIMIT datasets, in contrast to the FF architecture. Table \ref{tab:bench} shows SNN performance across various training methods for N-MNIST, DVS Gesture, and TIMIT classification. When trained with BPTT, the REC architecture exhibits a notable enhancement in accuracy when compared to the FF architecture. Similar accuracy improvements are also observed in the case of e-prop and DECOLLE. Benefits of recurrent weights are not evident in the N-MNIST classification task. Here, the performance of SNNs trained with BPTT and e-prop displays only slight differences between the REC module and the FF module. Interestingly, the explicit inclusion of recurrent weights leads to a performance degradation in SNNs trained with DECOLLE for the N-MNIST dataset. Further investigation is required to understand the underlying reason for this degradation.

\textbf{Additionally, a more local learning method slightly degrades the performance of SNNs, which indicates a trade-off between biological plausibility and performance.} In this study, the locality of the learning method is considered to increase from BPTT (global learning method) to e-prop and DECOLLE successively. In N-MNIST classification, a slight accuracy difference is observed between BPTT and e-prop in both REC and FF modules. DECOLLE exhibits an accuracy reduction in both architectures compared with e-prop.  A similar trend is observed for DVS Gesture and TIMIT classification tasks. In subsequent sections, we will focus on analyzing the performance differences and robustness between these methods. To provide a case study, we have selected the N-MNIST dataset.

\subsection{Representational Differences}
In the previous section, we observed a trade-off between biological plausibility and performance among various learning methods. Here, we investigate whether this performance discrepancy can be attributed to variations in the learned representations at corresponding layers across different training methods.

CKA is a technique used to measure the similarity between two sets of data, and it has been increasingly applied to understand the internal representation in neural networks. Pioneering work by Cortes et al. introduced CKA as an alignment measure for kernels in the context of support vector machines \cite{cortes2012algorithms}. Later, Kornblith et al. adapted CKA for deep learning, using it to compare layers across different neural network models, which provided insights into which layers of networks learn similar features \cite{ckann_kornblith2019similarity}. Williams et al. expanded this analysis to study the similarity between representations learned by convolutional neural networks \cite{williams2021generalized}. Li et al. investigated how SNNs differ from traditional ANNs in learning representations, using CKA to analyze both spatial and temporal dimensions \cite{ckasnn_li2023uncovering}. In this study, CKA is employed as a metric to assess the similarity of representations across layers that are learned using various methods with differing degrees of local learning. Consistent with prior works, the CKA is averaged over 4096 instances from the testing set of the N-MNIST dataset \cite{ckasnn_li2023uncovering, nguyen2020wide}. We also compare the similarity of learned representations of the various training methods with ANN-SNN conversion, where the SNN is trained using conventional ANN training methods \cite{sengupta2019going,lu2020exploring}.

\begin{figure}[t]
 \centerline{\includegraphics[width=0.48\textwidth]{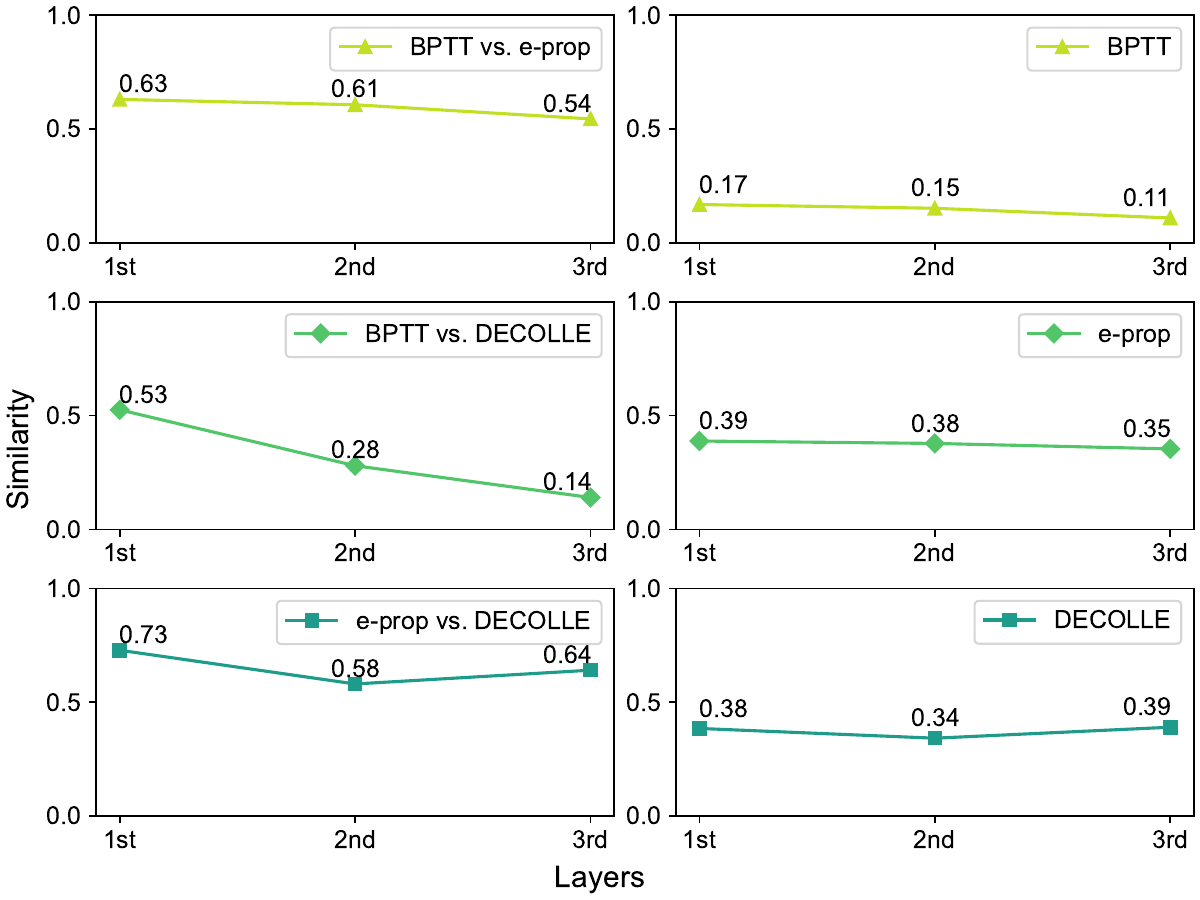}}
\caption{(Left) Layer-wise similarity among training methods; (Right) Similarity between ANN-SNN conversion method and different direct SNN training methods. The experiments are conducted on the N-MNIST dataset averaged over 4096 images from the test dataset. Significant differences exist between representations from ANN-SNN conversion and other methods. Additionally, training methods with closer locality typically show representational similarity between layers.}
\label{fig:cka}
\end{figure}

\textbf{Firstly, significant difference is observed between the representations learned from ANN-SNN conversion and those obtained through other training methods.} Figure \ref{fig:cka} plots the normalized HSIC between representations learned at each layer of SNN converted from ANN and that of SNNs trained directly. All the layer-wise similarities between the ANN-SNN conversion method and other direct SNN training methods are low, underscoring the fact that SNNs and ANNs learn very different representations.

\textbf{Secondly, aside from ANN-SNN conversion, we observe that methods with a closer level of locality exhibit a substantial degree of representational similarity between corresponding layers.} Figure \ref{fig:cka} plots the similarity between the representations of a layer trained by pairs of training methods. In particular, BPTT and e-prop, which have a closer level of locality, show a strong correlation in the representation of their first two layers based on CKA scores. E-prop and DECOLLE pair also exhibit a similar trend. However, the small representational similarity between corresponding layers trained by BPTT and DECOLLE is substantiated by a low CKA score, which addresses the performance gap between the two methods.

\subsection{Robustness under Attack}

In the previous section, we observed accuracy improvements after the addition of recurrence in spiking architectures, especially for sequential data processing. However, their impact on other aspects of SNN performance, like robustness to adversarial attacks, remains unclear. In this section, we explore this aspect along with empirical tools to explain our observations.

\begin{table}[t]
\caption{NN accuracy (\%) under FGSM attack is evaluated on the N-MNIST dataset. SNNs are trained by methods with varying localities. Model robustness improves with explicit recurrent weights as well as increasing locality of training methods.}
\begin{center}
\begin{tabular}{|l|lllll|}
\hline
\textbf{Epsilon}                                &  \multicolumn{1}{l|}{\textbf{0.001}} & \multicolumn{1}{l|}{\textbf{0.005}} & \multicolumn{1}{l|}{\textbf{0.01}} & \multicolumn{1}{l|}{\textbf{0.02}} & \multicolumn{1}{l|}{\textbf{0.05}} \\ 
\hline
\multicolumn{1}{|c|}{\textbf{Training Methods}} & \multicolumn{5}{c|}{\textit{FF}}                                              \\ \hline
BPTT                                             & \multicolumn{1}{l|}{83.82}        & \multicolumn{1}{l|}{16.78}        & \multicolumn{1}{l|}{4.34}        & \multicolumn{1}{l|}{1.22}        & \multicolumn{1}{l|}{0.73}              \\ \hline
e-prop                                            & \multicolumn{1}{l|}{89.90}        & \multicolumn{1}{l|}{41.27}        & \multicolumn{1}{l|}{12.33}        & \multicolumn{1}{l|}{3.27}        & \multicolumn{1}{l|}{2.87}            \\ \hline
DECOLLE                                          & \multicolumn{1}{l|}{83.48}        & \multicolumn{1}{l|}{45.11}        & \multicolumn{1}{l|}{28.57}       & \multicolumn{1}{l|}{21.29}       & \multicolumn{1}{l|}{17.79}          \\ \hline
\multicolumn{1}{|c|}{\textbf{Training Methods}} & \multicolumn{5}{c|}{\textit{REC}} \\ \hline
BPTT                                              & \multicolumn{1}{l|}{94.66}        & \multicolumn{1}{l|}{80.76}        & \multicolumn{1}{l|}{58.73}       & \multicolumn{1}{l|}{27.69}        & \multicolumn{1}{l|}{18.55}     \\ \hline
e-prop                                             & \multicolumn{1}{l|}{93.04}        & \multicolumn{1}{l|}{78.28}        & \multicolumn{1}{l|}{60.57}       & \multicolumn{1}{l|}{35.07}       & \multicolumn{1}{l|}{18.25}           \\ \hline
DECOLLE                                           & \multicolumn{1}{l|}{86.76}        & \multicolumn{1}{l|}{84.14}        & \multicolumn{1}{l|}{72.81}       & \multicolumn{1}{l|}{47.56}       & \multicolumn{1}{l|}{23.75}            \\ \hline

\end{tabular}
\label{tab:fgsm}
\end{center}
\end{table}

\begin{table*}[t]
\caption{Attack success rate (ASR) of targeted backdoor attack on SNNs along with their accuracy under attack is measured on the N-MNIST dataset with 5 random selections of target and source classes. Local training methods do not demonstrate robustness against backdoor attacks.}
\begin{center}
\begin{tabular}{|lcccccccccc|}
\hline
\multicolumn{1}{|l|}{\textbf{Poison Rate}}      & \multicolumn{2}{c|}{\textbf{0.1}}                                     & \multicolumn{2}{c|}{\textbf{0.3}}                                     & \multicolumn{2}{c|}{\textbf{0.5}}                                     & \multicolumn{2}{c|}{\textbf{0.7}}                                     & \multicolumn{2}{c|}{\textbf{0.9}}                                     \\ \hline
\multicolumn{11}{|c|}{\textit{FF}}                                                       \\ \hline
\multicolumn{1}{|l|}{\textbf{Training Methods}} & \multicolumn{1}{l|}{\textbf{Accuracy}} & \multicolumn{1}{l|}{\textbf{ASR}} & \multicolumn{1}{l|}{\textbf{Accuracy}} & \multicolumn{1}{l|}{\textbf{ASR}} & \multicolumn{1}{l|}{\textbf{Accuracy}} & \multicolumn{1}{l|}{\textbf{ASR}} & \multicolumn{1}{l|}{\textbf{Accuracy}} & \multicolumn{1}{l|}{\textbf{ASR}} & \multicolumn{1}{l|}{\textbf{Accuracy}} & \multicolumn{1}{l|}{\textbf{ASR}} \\ \hline
\multicolumn{1}{|l|}{BPTT}             & \multicolumn{1}{c|}{96.23\%}   & \multicolumn{1}{c|}{4.80\%}  & \multicolumn{1}{c|}{93.34\%}   & \multicolumn{1}{c|}{33.54\%}  & \multicolumn{1}{c|}{91.62\%}   & \multicolumn{1}{c|}{43.76\%}  & \multicolumn{1}{c|}{89.64\%}   & \multicolumn{1}{c|}{66.06\%}  & \multicolumn{1}{c|}{87.47\%}   & 78.19\%                       \\ \hline
\multicolumn{1}{|l|}{e-prop}            & \multicolumn{1}{c|}{96.85\%}   & \multicolumn{1}{c|}{1.93\%}  & \multicolumn{1}{c|}{93.46\%}   & \multicolumn{1}{c|}{33.35\%}  & \multicolumn{1}{c|}{92.30\%}   & \multicolumn{1}{c|}{46.94\%}  & \multicolumn{1}{c|}{89.38\%}   & \multicolumn{1}{c|}{75.09\%}  & \multicolumn{1}{c|}{87.32\%}   & 96.95\%                       \\ \hline
\multicolumn{1}{|l|}{DECOLLE}          & \multicolumn{1}{c|}{92.74\%}   & \multicolumn{1}{c|}{4.66\%}  & \multicolumn{1}{c|}{89.88\%}   & \multicolumn{1}{c|}{30.04\%}  & \multicolumn{1}{c|}{88.14\%}   & \multicolumn{1}{c|}{48.31\%}  & \multicolumn{1}{c|}{85.12\%}   & \multicolumn{1}{c|}{76.77\%}  & \multicolumn{1}{c|}{83.62\%}   & 90.80\%                       \\ \hline
\multicolumn{11}{|c|}{\textit{REC}}                                                  \\ \hline
\multicolumn{1}{|l|}{\textbf{Training Methods}} & \multicolumn{1}{l|}{\textbf{Accuracy}} & \multicolumn{1}{l|}{\textbf{ASR}} & \multicolumn{1}{l|}{\textbf{Accuracy}} & \multicolumn{1}{l|}{\textbf{ASR}} & \multicolumn{1}{l|}{\textbf{Accuracy}} & \multicolumn{1}{l|}{\textbf{ASR}} & \multicolumn{1}{l|}{\textbf{Accuracy}} & \multicolumn{1}{l|}{\textbf{ASR}} & \multicolumn{1}{l|}{\textbf{Accuracy}} & \multicolumn{1}{l|}{\textbf{ASR}} \\ \hline
\multicolumn{1}{|l|}{BPTT}             & \multicolumn{1}{c|}{96.43\%}   & \multicolumn{1}{c|}{2.64\%}  & \multicolumn{1}{c|}{94.19\%}   & \multicolumn{1}{c|}{22.03\%}  & \multicolumn{1}{c|}{92.24\%}   & \multicolumn{1}{c|}{43.78\%}  & \multicolumn{1}{c|}{91.22\%}   & \multicolumn{1}{c|}{62.24\%}  & \multicolumn{1}{c|}{87.38\%}   & 90.73\%                       \\ \hline
\multicolumn{1}{|l|}{e-prop}            & \multicolumn{1}{c|}{96.69\%}   & \multicolumn{1}{c|}{0.45\%}  & \multicolumn{1}{c|}{94.81\%}   & \multicolumn{1}{c|}{17.65\%}  & \multicolumn{1}{c|}{91.92\%}   & \multicolumn{1}{c|}{44.56\%}  & \multicolumn{1}{c|}{88.51\%}   & \multicolumn{1}{c|}{74.15\%}  & \multicolumn{1}{c|}{87.23\%}   & 93.15\%                       \\ \hline
\multicolumn{1}{|l|}{DECOLLE}          & \multicolumn{1}{c|}{90.21\%}   & \multicolumn{1}{c|}{0.86\%}  & \multicolumn{1}{c|}{87.46\%}   & \multicolumn{1}{c|}{23.75\%}  & \multicolumn{1}{c|}{81.55\%}   & \multicolumn{1}{c|}{74.10\%}  & \multicolumn{1}{c|}{81.29\%}   & \multicolumn{1}{c|}{88.96\%}  & \multicolumn{1}{c|}{80.81\%}   & 91.02\%                      \\ \hline

\end{tabular}
\label{tab:backdoor}
\end{center}
\end{table*}

Specifically, the Fast Gradient Sign Method (FGSM) attack leverages the sign of gradients to create adversarial instances that maximize the loss of a neural network, misleading it to make incorrect classifications \cite{fgsm_goodfellow2014explaining}. However, due to the non-differentiable nature of SNNs, the direct calculation of gradients is not applicable. To solve this problem, previous work has proposed an effective framework which involves crafting an adversarial example for an SNN by creating its ANN counterpart and then calculating gradients for the corresponding ANN \cite{fgsm2_sharmin2019comprehensive}. These gradients are subsequently utilized to construct an adversarial example for the SNN. We adopt this variation of the FGSM attack to compare the robustness of various training methods and to investigate the impact of recurrent weights on robustness.

Then, we employ a targeted backdoor attack, which is a special type of poisoning attack. Such an attack involves the process of injecting a malicious imperceptible trigger during the training phase. Following the trigger mechanism established in previous work \cite{bd_badnet_gu2019badnets}, we initiate the poisoning of images by modifying 4 pixel locations at all time steps in the input data. For each poison ratio, five random sources and targets are selected. To provide a more robust and unbiased assessment, we calculate the average results from 5 runs with randomly assigned sources and targets, thus mitigating the potential bias introduced by specific poisoning choices. To assess the impact of these backdoor attacks, we examine changes in accuracy and attack success rate (ASR), which is the proportion of labels for the source class that are predicted as the target class. A lower ASR signifies greater robustness.

Table \ref{tab:fgsm} reports the performance of different training methods when subjected to FGSM attacks, where the perturbation level, denoted as ``Epsilon'', varies across a range from 0.001 to 0.05. The results of our benchmarking analysis for backdoor attacks are illustrated in Table \ref{tab:backdoor}. We mainly focus on poison rates in realistic settings below 50\% as shown in prior works \cite{bd1_chen2017targeted, bd2_adi2018turning, bd3_gao2019strip, bd_badnet_gu2019badnets}. Both attacks are evaluated on the N-MNIST dataset.

\textbf{Enhancement of the model's robustness is observed with the inclusion of explicit recurrent weights.}
Under FGSM attacks, compared to FF models trained by the same methods and subjected to the same ``Epsilon'' values, the REC model retains more accuracy than the FF model. Besides, the robustness advantage of explicit recurrence is also observed under backdoor attack. Specifically, under more realistic settings with poison rates at 10\% and 30\%, we observe that the REC model consistently exhibits a lower ASR compared to its corresponding FF model in all three training methods.

\textbf{Local learning methods showed improved robustness under FGSM attacks compared to global learning methods.}
Specifically, the local learning method DECOLLE demonstrates enhanced robustness against attacks with high perturbation values. In contrast, the global training methods experience a significant drop in accuracy under high ``Epsilon'' values, which is consistent in both FF and REC architectures. This advantage can be attributed to the local training methods' use of individual loss functions for each layer, which enhances robustness in comparison to global loss functions that are vulnerable to globally calculated gradients. 
\textbf{Under backdoor attacks, where global gradients are not used, such robustness gains of local learning methods are not observed.} In the previous section, we show that there is a trade-off between biological plausibility and performance. Such a trade-off is not demonstrated for adversarial attacks. Local learning methods retain robustness at par with global learning methods with respect to non-gradient-based attacks and exhibit enhanced robustness for gradient-based attacks. 

\subsection{CKA Measurements to Explain Robustness\label{sec:cka_rob}}

\begin{figure}[htbp]
\centerline{\includegraphics[width=0.48\textwidth]{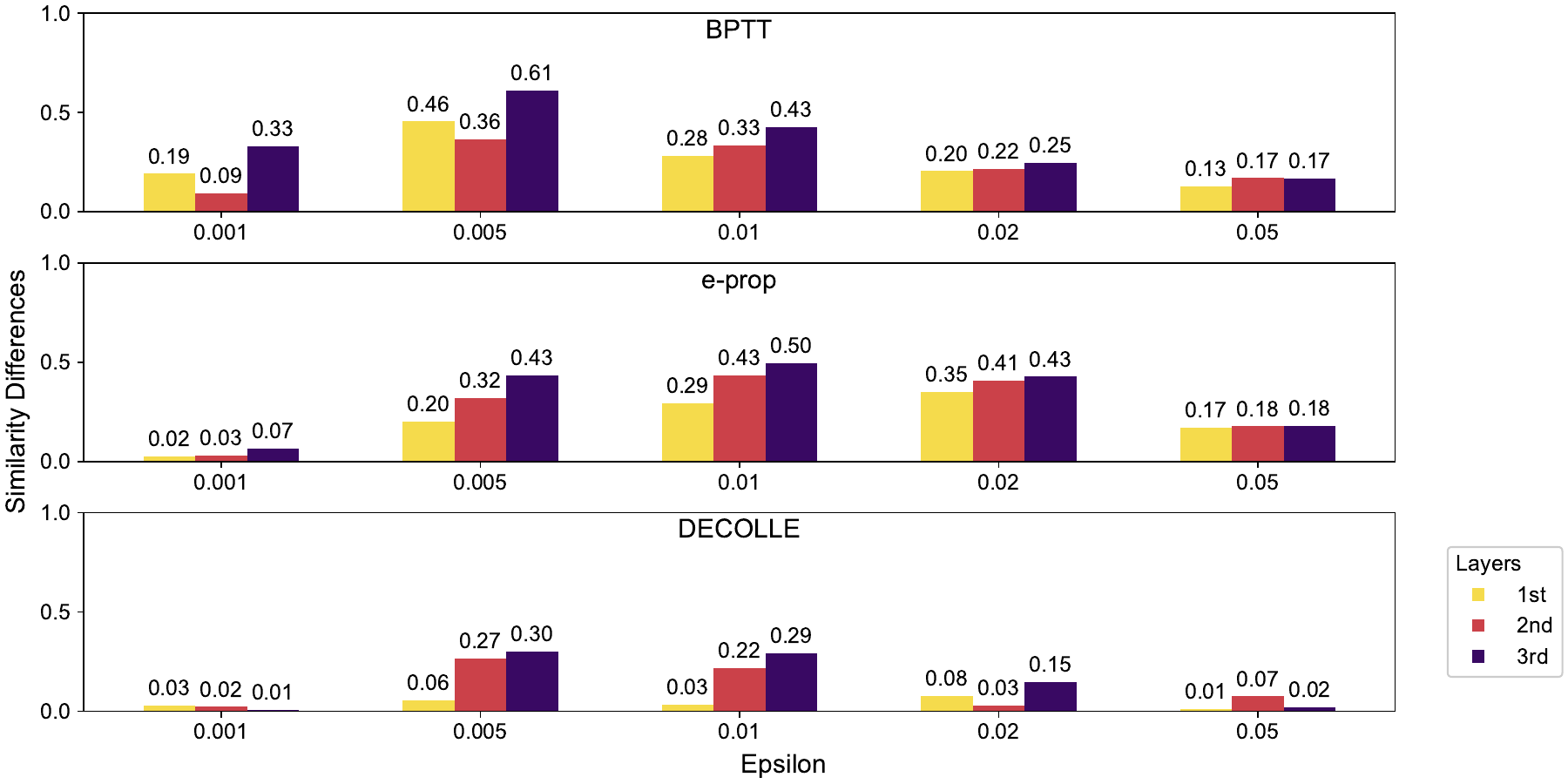}}
\caption{Layerwise representational similarity between clean and adversarial input data of FF architecture subtracted from the REC architecture for FGSM adversarial attacks. The experiments are conducted on the N-MNIST dataset. Adding recurrent weights improves the robustness leading to higher similarity between clean and adversarial inputs. Robustness improvements due to recurrent weights are more pronounced in global methods.}
\label{fig:cka:robust}
\end{figure}

In this section, we further assess the robustness of SNNs from the perspective of CKA. Prior research has quantified the robustness of SNNs directly trained under adversarial attacks \cite{ckasnn_li2023uncovering} and established a correlation between robustness against adversarial attacks and the degree of representational similarity of a particular network between clean and perturbed input images \cite{madry2017towards}. Here, we employ the CKA metric to evaluate the robustness of SNNs trained using each training method under FGSM attacks with the same set of ``Epsilon'' values used in the previous section. Figure \ref{fig:cka:robust} measures the layerwise representational similarity between clean and adversarial inputs of REC architecture subtracted by the similarity of FF architecture under FGSM attacks. Each similarity is computed by contrasting representations of a layer by passing clean data and a layer at the same architectural level by passing adversarial data. Each color corresponds to one specific layer.

\textbf{It is observed that the addition of recurrent weights contributes to better preservation of the representations learned at each layer across these training methods.} In these experiments, positive similarity differences are obtained at all perturbation levels, revealing that REC SNNs maintain larger representational similarity (implying more robust representations learned in the presence of attacks) compared to FF SNNs between clean and adversarial data. \textbf{The results also reveal that global learning methods exhibit a greater enhancement in robustness with recurrent connections}, as the similarity differences are larger in BPTT than in e-prop and DECOLLE. The trend matches the data provided in Table \ref{tab:fgsm}. Through this analysis, we provide evidence that the inclusion of recurrent weights enhances the robustness of neuromorphic architectures. 

\subsection{Relative Importance of Recurrent Weights \label{sec:fisher}}

In the previous section, we demonstrated that there is a robustness advantage for explicit recurrent SNNs. Here, we examine the importance of additional recurrence in SNNs from the perspective of Fisher information. A high Fisher information exhibited in weight connections is shown to play a substantial role in making predictions for the given data \cite{achille2018critical, kirkpatrick2017overcoming}. Specifically, the Fisher information in SNNs evaluates the accumulation of temporal information during the training process. In previous work \cite{fisher_kim2023exploring}, normalized layer-wise Fisher information was calculated across time steps, providing an assessment of the relative importance of each layer. We follow the same procedure to calculate Fisher information during the training process by capturing the gradient of weights.

\begin{figure}[htbp]
\centerline{\includegraphics[width=0.48\textwidth]{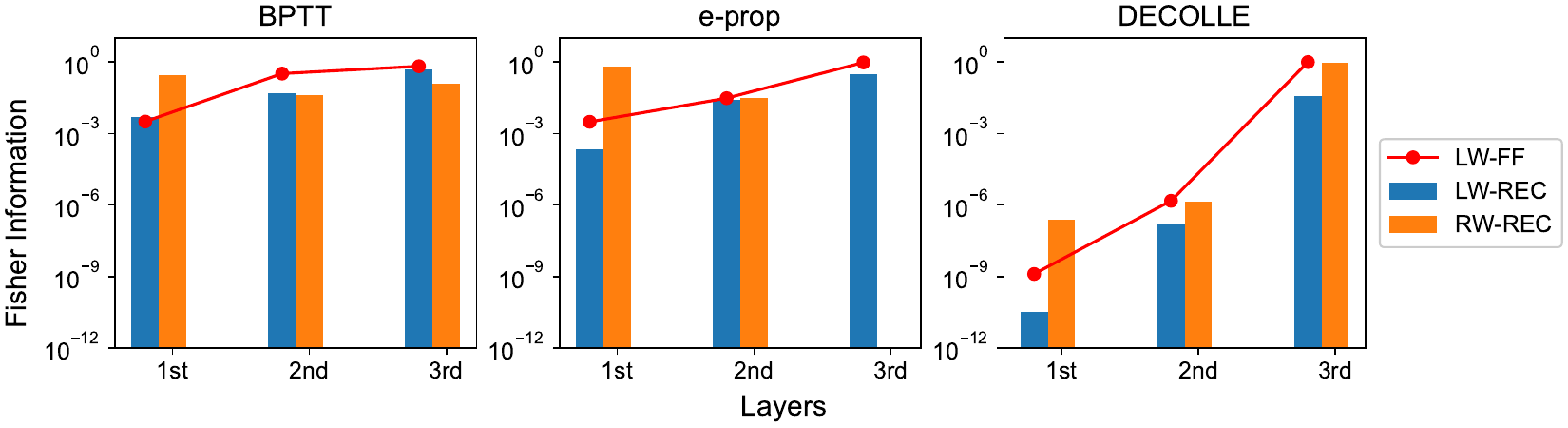}}
\caption{Layerwise Fisher information measured in SNNs trained by learning methods with varying degrees of locality on the N-MNIST dataset averaged over 5 independent runs. Recurrent weights are as important as linear weights, and layerwise increments in linear weights' importance is independent of recurrent weights.}
\label{fig:fisher}
\end{figure}

In Figure \ref{fig:fisher}, we present the Fisher information of linear weights in the FF architecture, denoted as LW-FF in the figure. We then compare this Fisher information with both linear and recurrent weights in the REC architecture, denoted as LW-REC and RW-REC in the graph. It is important to note that due to the nature of e-prop, the additional recurrent weights are not added to the last layer in the REC architecture. 

\textbf{Recurrent weights are demonstrated to have comparable importance with respect to linear weights.} After the addition of explicit recurrent weights, the importance of linear weights (LW-REC) degrades in comparison to linear weights in the FF model (LW-FF). On the other hand, the importance of recurrent weights (RW-REC) shows comparable Fisher information values with regards to linear weights (LW-REC). Based on these two observations, we can conclude that the degradation of importance in linear weights from the values in the FF model (LW-FF) is due to reallocation of importance to linear weights (LW-REC) and recurrent weights (RW-REC) in the REC model since Fisher information is normalized such that the summation of values of all layers is one.

\textbf{Additionally, it is found that layerwise increments in the relative importance of linear weights are independent of recurrent weights.}
Specifically, an increase of relative importance from input layer to output layer is observed among all training methods. Although the magnitude of this trend may vary across different methods, this trend remains consistent independent of the addition of extra recurrent weights (LW-REC). The magnitude variations mainly come from the emphasis on input layers. In particular, as the locality of learning methods shifts from global to local, the importance of the input layer drops significantly.

\section{Conclusions}
Edge devices like smartphones, Internet of things (IoT) gadgets, and embedded systems require improved energy efficiency due to strict memory and computational constraints \cite{covi2021adaptive, gao2023presynaptic}. Recent advancements in local learning methods enable on-chip learning \cite{pehle2022brainscales, frenkel2022reckon}, allowing models to be trained and updated directly on edge devices with significantly reduced energy consumption compared to global learning methods \cite{liu2018memory, eprop_bellec2020solution, decolle_kaiser2020synaptic}.
In this work, these local learning methods are compared in terms of performance and robustness. To understand the performance disparity of these methods, representational differences are examined through the CKA metric. SNN learning methods with a higher degree of locality are shown to have performance degradation, but they show better robustness under FGSM adversarial attacks.
However, the robustness advantages of local learning methods are not observed for non-gradient-based attacks like backdoor attacks. Additionally, we investigate the contribution of additional recurrent weights on the performance of SNNs. Measurements on the relative importance of recurrent weights are performed using Fisher information. We illustrate that recurrent weights not only have positive influences on the performance of SNNs but also augment the robustness of SNNs. These observations are further substantiated by the fact that recurrent weights and linear weights exhibit comparable informative significance. 
Overall, this work provides valuable insights into the selection of learning methods for SNNs and offers a benchmarking framework for future SNN designers. We acknowledge that our conclusions may currently be limited to feed-forward architectures. 
Our observations could potentially be generalized to more complex network topologies. 
Additionally, the integration of more advanced analytical tools could further enhance these insights. Future work could also focus on extending our analysis to convolutional layers \cite{fang2021deep, hu2021spiking} and transformer-based SNN models \cite{zhou2022spikformer, zhang2022spiking, bal2024spikingbert}.
\section*{Acknowledgments}
This material is based upon work supported in part by the U.S. Department of Energy, Office of Science, Office of Advanced Scientific Computing Research, under Award Number \#DE-SC0021562, the U.S. National Science Foundation under award No. CCF \#1955815, CAREER \# 2337646 and EFRI BRAID \#2318101 and by Oracle Cloud credits and related resources provided by the Oracle for Research
program.
\vspace{12pt}


\begin{thebibliography}{10}
\providecommand{\url}[1]{#1}
\csname url@samestyle\endcsname
\providecommand{\newblock}{\relax}
\providecommand{\bibinfo}[2]{#2}
\providecommand{\BIBentrySTDinterwordspacing}{\spaceskip=0pt\relax}
\providecommand{\BIBentryALTinterwordstretchfactor}{4}
\providecommand{\BIBentryALTinterwordspacing}{\spaceskip=\fontdimen2\font plus
\BIBentryALTinterwordstretchfactor\fontdimen3\font minus
  \fontdimen4\font\relax}
\providecommand{\BIBforeignlanguage}[2]{{%
\expandafter\ifx\csname l@#1\endcsname\relax
\typeout{** WARNING: IEEEtran.bst: No hyphenation pattern has been}%
\typeout{** loaded for the language `#1'. Using the pattern for}%
\typeout{** the default language instead.}%
\else
\language=\csname l@#1\endcsname
\fi
#2}}
\providecommand{\BIBdecl}{\relax}
\BIBdecl
\bibitem{maass1997networks}
W.~Maass, ``Networks of spiking neurons: the third generation of neural network models,'' \emph{Neural networks}, vol.~10, no.~9, pp. 1659--1671, 1997.

\bibitem{gerstner2002spiking}
W.~Gerstner and W.~M. Kistler, \emph{Spiking neuron models: Single neurons, populations, plasticity}.\hskip 1em plus 0.5em minus 0.4em\relax Cambridge university press, 2002.

\bibitem{tavanaei2019deep}
A.~Tavanaei, M.~Ghodrati, S.~R. Kheradpisheh, T.~Masquelier, and A.~Maida, ``Deep learning in spiking neural networks,'' \emph{Neural networks}, vol. 111, pp. 47--63, 2019.

\bibitem{diehl2015unsupervised}
P.~U. Diehl and M.~Cook, ``Unsupervised learning of digit recognition using spike-timing-dependent plasticity,'' \emph{Frontiers in computational neuroscience}, vol.~9, p.~99, 2015.

\bibitem{bi1998synaptic}
G.-q. Bi and M.-m. Poo, ``Synaptic modifications in cultured hippocampal neurons: dependence on spike timing, synaptic strength, and postsynaptic cell type,'' \emph{Journal of neuroscience}, vol.~18, no.~24, pp. 10\,464--10\,472, 1998.

\bibitem{mead1990neuromorphic}
C.~Mead, ``Neuromorphic electronic systems,'' \emph{Proceedings of the IEEE}, vol.~78, no.~10, pp. 1629--1636, 1990.

\bibitem{merolla2014million}
P.~A. Merolla, J.~V. Arthur, R.~Alvarez-Icaza, A.~S. Cassidy, J.~Sawada, F.~Akopyan, B.~L. Jackson, N.~Imam, C.~Guo, Y.~Nakamura \emph{et~al.}, ``A million spiking-neuron integrated circuit with a scalable communication network and interface,'' \emph{Science}, vol. 345, no. 6197, pp. 668--673, 2014.

\bibitem{benjamin2014neurogrid}
B.~V. Benjamin, P.~Gao, E.~McQuinn, S.~Choudhary, A.~R. Chandrasekaran, J.-M. Bussat, R.~Alvarez-Icaza, J.~V. Arthur, P.~A. Merolla, and K.~Boahen, ``Neurogrid: A mixed-analog-digital multichip system for large-scale neural simulations,'' \emph{Proceedings of the IEEE}, vol. 102, no.~5, pp. 699--716, 2014.

\bibitem{boahen2006neuromorphic}
K.~Boahen, ``Neuromorphic microchips,'' \emph{Scientific American}, vol.~16, no.~3, pp. 20--27, 2006.

\bibitem{davies2018loihi}
M.~Davies, N.~Srinivasa, T.-H. Lin, G.~Chinya, Y.~Cao, S.~H. Choday, G.~Dimou, P.~Joshi, N.~Imam, S.~Jain \emph{et~al.}, ``Loihi: A neuromorphic manycore processor with on-chip learning,'' \emph{Ieee Micro}, vol.~38, no.~1, pp. 82--99, 2018.

\bibitem{furber2014spinnaker}
S.~B. Furber, F.~Galluppi, S.~Temple, and L.~A. Plana, ``The spinnaker project,'' \emph{Proceedings of the IEEE}, vol. 102, no.~5, pp. 652--665, 2014.

\bibitem{ponghiran2022spiking}
W.~Ponghiran and K.~Roy, ``Spiking neural networks with improved inherent recurrence dynamics for sequential learning,'' in \emph{Proceedings of the AAAI Conference on Artificial Intelligence}, vol.~36, no.~7, 2022, pp. 8001--8008.

\bibitem{lsnn_bellec2018long}
G.~Bellec, D.~Salaj, A.~Subramoney, R.~Legenstein, and W.~Maass, ``Long short-term memory and learning-to-learn in networks of spiking neurons,'' \emph{Advances in neural information processing systems}, vol.~31, 2018.

\bibitem{yin2021accurate}
B.~Yin, F.~Corradi, and S.~M. Boht{\'e}, ``Accurate and efficient time-domain classification with adaptive spiking recurrent neural networks,'' \emph{Nature Machine Intelligence}, vol.~3, no.~10, pp. 905--913, 2021.

\bibitem{eprop_bellec2020solution}
G.~Bellec, F.~Scherr, A.~Subramoney, E.~Hajek, D.~Salaj, R.~Legenstein, and W.~Maass, ``A solution to the learning dilemma for recurrent networks of spiking neurons,'' \emph{Nature communications}, vol.~11, no.~1, p. 3625, 2020.

\bibitem{bal2023spikingbert}
M.~Bal and A.~Sengupta, ``Spikingbert: Distilling bert to train spiking language models using implicit differentiation,'' \emph{arXiv preprint arXiv:2308.10873}, 2023.

\bibitem{lillicrap2016random}
T.~P. Lillicrap, D.~Cownden, D.~B. Tweed, and C.~J. Akerman, ``Random synaptic feedback weights support error backpropagation for deep learning,'' \emph{Nature communications}, vol.~7, no.~1, p. 13276, 2016.

\bibitem{baldi2017learning}
P.~Baldi, P.~Sadowski, and Z.~Lu, ``Learning in the machine: the symmetries of the deep learning channel,'' \emph{Neural Networks}, vol.~95, pp. 110--133, 2017.

\bibitem{decolle_kaiser2020synaptic}
J.~Kaiser, H.~Mostafa, and E.~Neftci, ``Synaptic plasticity dynamics for deep continuous local learning (decolle),'' \emph{Frontiers in Neuroscience}, vol.~14, p. 424, 2020.

\bibitem{crick1989recent}
F.~Crick, ``The recent excitement about neural networks,'' \emph{Nature}, vol. 337, pp. 129--132, 1989.

\bibitem{lillicrap2020backpropagation}
T.~P. Lillicrap, A.~Santoro, L.~Marris, C.~J. Akerman, and G.~Hinton, ``Backpropagation and the brain,'' \emph{Nature Reviews Neuroscience}, vol.~21, no.~6, pp. 335--346, 2020.

\bibitem{werbos1990backpropagation}
P.~J. Werbos, ``Backpropagation through time: what it does and how to do it,'' \emph{Proceedings of the IEEE}, vol.~78, no.~10, pp. 1550--1560, 1990.

\bibitem{williams1995gradient}
R.~J. Williams and D.~Zipser, \emph{Gradient-Based Learning Algorithms for Recurrent Networks and Their Computational Complexity}.\hskip 1em plus 0.5em minus 0.4em\relax USA: L. Erlbaum Associates Inc., 1995, p. 433–486.

\bibitem{lu2022neuroevolution}
S.~Lu and A.~Sengupta, ``Neuroevolution guided hybrid spiking neural network training,'' \emph{Frontiers in neuroscience}, vol.~16, p. 838523, 2022.

\bibitem{zenke2018superspike}
F.~Zenke and S.~Ganguli, ``Superspike: Supervised learning in multilayer spiking neural networks,'' \emph{Neural computation}, vol.~30, no.~6, pp. 1514--1541, 2018.

\bibitem{neftci2019surrogate}
E.~O. Neftci, H.~Mostafa, and F.~Zenke, ``Surrogate gradient learning in spiking neural networks: Bringing the power of gradient-based optimization to spiking neural networks,'' \emph{IEEE Signal Processing Magazine}, vol.~36, no.~6, pp. 51--63, 2019.

\bibitem{bal2022sequence}
M.~Bal and A.~Sengupta, ``Sequence learning using equilibrium propagation,'' \emph{arXiv preprint arXiv:2209.09626}, 2022.

\bibitem{deng2020rethinking}
L.~Deng, Y.~Wu, X.~Hu, L.~Liang, Y.~Ding, G.~Li, G.~Zhao, P.~Li, and Y.~Xie, ``Rethinking the performance comparison between snns and anns,'' \emph{Neural networks}, vol. 121, pp. 294--307, 2020.

\bibitem{ckasnn_li2023uncovering}
Y.~Li, Y.~Kim, H.~Park, and P.~Panda, ``Uncovering the representation of spiking neural networks trained with surrogate gradient,'' \emph{arXiv preprint arXiv:2304.13098}, 2023.

\bibitem{he2020comparing}
W.~He, Y.~Wu, L.~Deng, G.~Li, H.~Wang, Y.~Tian, W.~Ding, W.~Wang, and Y.~Xie, ``Comparing snns and rnns on neuromorphic vision datasets: Similarities and differences,'' \emph{Neural Networks}, vol. 132, pp. 108--120, 2020.

\bibitem{sharmin2019comprehensive}
S.~Sharmin, P.~Panda, S.~S. Sarwar, C.~Lee, W.~Ponghiran, and K.~Roy, ``A comprehensive analysis on adversarial robustness of spiking neural networks,'' in \emph{2019 IJCNN}.\hskip 1em plus 0.5em minus 0.4em\relax IEEE, 2019, pp. 1--8.

\bibitem{liang2021exploring}
L.~Liang, X.~Hu, L.~Deng, Y.~Wu, G.~Li, Y.~Ding, P.~Li, and Y.~Xie, ``Exploring adversarial attack in spiking neural networks with spike-compatible gradient,'' \emph{IEEE transactions on neural networks and learning systems}, 2021.

\bibitem{marchisio2020spiking}
A.~Marchisio, G.~Nanfa, F.~Khalid, M.~A. Hanif, M.~Martina, and M.~Shafique, ``Is spiking secure? a comparative study on the security vulnerabilities of spiking and deep neural networks,'' in \emph{2020 International Joint Conference on Neural Networks (IJCNN)}.\hskip 1em plus 0.5em minus 0.4em\relax IEEE, 2020, pp. 1--8.

\bibitem{bd_badnet_gu2019badnets}
T.~Gu, B.~Dolan-Gavitt, and S.~Garg, ``Badnets: Identifying vulnerabilities in the machine learning model supply chain.(2017),'' \emph{arXiv preprint arXiv:1708.06733}, 2019.

\bibitem{bhargava2022gradient}
A.~Bhargava, M.~R. Rezaei, and M.~Lankarany, ``Gradient-free neural network training via synaptic-level reinforcement learning,'' \emph{AppliedMath}, vol.~2, no.~2, pp. 185--195, 2022.

\bibitem{kundu2021hire}
S.~Kundu, M.~Pedram, and P.~A. Beerel, ``Hire-snn: Harnessing the inherent robustness of energy-efficient deep spiking neural networks by training with crafted input noise,'' in \emph{Proceedings of the IEEE/CVF International Conference on Computer Vision}, 2021, pp. 5209--5218.

\bibitem{ding2022snn}
J.~Ding, T.~Bu, Z.~Yu, T.~Huang, and J.~Liu, ``Snn-rat: Robustness-enhanced spiking neural network through regularized adversarial training,'' \emph{Advances in Neural Information Processing Systems}, vol.~35, pp. 24\,780--24\,793, 2022.

\bibitem{ozdenizci2023adversarially}
O.~{\"O}zdenizci and R.~Legenstein, ``Adversarially robust spiking neural networks through conversion,'' \emph{arXiv preprint arXiv:2311.09266}, 2023.

\bibitem{rumelhart1986learning}
D.~E. Rumelhart, G.~E. Hinton, and R.~J. Williams, ``Learning representations by back-propagating errors,'' \emph{nature}, vol. 323, no. 6088, pp. 533--536, 1986.

\bibitem{EligibilityTraces}
\BIBentryALTinterwordspacing
W.~Gerstner, M.~Lehmann, V.~Liakoni, D.~Corneil, and J.~Brea, ``Eligibility traces and plasticity on behavioral time scales: Experimental support of neohebbian three-factor learning rules,'' \emph{Frontiers in Neural Circuits}, vol.~12, 2018. [Online]. Available: \url{https://www.frontiersin.org/articles/10.3389/fncir.2018.00053}
\BIBentrySTDinterwordspacing

\bibitem{fisher1925theory}
R.~A. Fisher, ``Theory of statistical estimation,'' in \emph{Mathematical proceedings of the Cambridge philosophical society}, vol.~22, no.~5.\hskip 1em plus 0.5em minus 0.4em\relax Cambridge University Press, 1925, pp. 700--725.

\bibitem{fisher_kim2023exploring}
Y.~Kim, Y.~Li, H.~Park, Y.~Venkatesha, A.~Hambitzer, and P.~Panda, ``Exploring temporal information dynamics in spiking neural networks,'' in \emph{Proceedings of the AAAI Conference on Artificial Intelligence}, vol.~37, no.~7, 2023, pp. 8308--8316.

\bibitem{ckann_kornblith2019similarity}
S.~Kornblith, M.~Norouzi, H.~Lee, and G.~Hinton, ``Similarity of neural network representations revisited,'' in \emph{International conference on machine learning}.\hskip 1em plus 0.5em minus 0.4em\relax PMLR, 2019, pp. 3519--3529.

\bibitem{greenfeld2020robust}
D.~Greenfeld and U.~Shalit, ``Robust learning with the hilbert-schmidt independence criterion,'' in \emph{International Conference on Machine Learning}.\hskip 1em plus 0.5em minus 0.4em\relax PMLR, 2020, pp. 3759--3768.

\bibitem{nguyen2020wide}
T.~Nguyen, M.~Raghu, and S.~Kornblith, ``Do wide and deep networks learn the same things? uncovering how neural network representations vary with width and depth,'' \emph{arXiv preprint arXiv:2010.15327}, 2020.

\bibitem{song2012feature}
L.~Song, A.~Smola, A.~Gretton, J.~Bedo, and K.~Borgwardt, ``Feature selection via dependence maximization.'' \emph{Journal of Machine Learning Research}, vol.~13, no.~5, 2012.

\bibitem{orchard2015NMNIST}
G.~Orchard, A.~Jayawant, G.~K. Cohen, and N.~Thakor, ``Converting static image datasets to spiking neuromorphic datasets using saccades,'' \emph{Frontiers in neuroscience}, vol.~9, p. 437, 2015.

\bibitem{deng2012mnist}
L.~Deng, ``The mnist database of handwritten digit images for machine learning research [best of the web],'' \emph{IEEE signal processing magazine}, vol.~29, no.~6, pp. 141--142, 2012.

\bibitem{lee2016training}
J.~H. Lee, T.~Delbruck, and M.~Pfeiffer, ``Training deep spiking neural networks using backpropagation,'' \emph{Frontiers in neuroscience}, vol.~10, p. 508, 2016.

\bibitem{amir2017low}
A.~Amir, B.~Taba, D.~Berg, T.~Melano, J.~McKinstry, C.~Di~Nolfo, T.~Nayak, A.~Andreopoulos, G.~Garreau, M.~Mendoza \emph{et~al.}, ``A low power, fully event-based gesture recognition system,'' in \emph{Proceedings of the IEEE conference on computer vision and pattern recognition}, 2017, pp. 7243--7252.

\bibitem{garofolo1993darpa}
J.~S. Garofolo, L.~F. Lamel, W.~M. Fisher, J.~G. Fiscus, and D.~S. Pallett, ``Darpa timit acoustic-phonetic continous speech corpus cd-rom. nist speech disc 1-1.1,'' \emph{NASA STI/Recon technical report n}, vol.~93, p. 27403, 1993.

\bibitem{cortes2012algorithms}
C.~Cortes, M.~Mohri, and A.~Rostamizadeh, ``Algorithms for learning kernels based on centered alignment,'' \emph{The Journal of Machine Learning Research}, vol.~13, no.~1, pp. 795--828, 2012.

\bibitem{williams2021generalized}
A.~H. Williams, E.~Kunz, S.~Kornblith, and S.~Linderman, ``Generalized shape metrics on neural representations,'' \emph{Advances in Neural Information Processing Systems}, vol.~34, pp. 4738--4750, 2021.

\bibitem{sengupta2019going}
A.~Sengupta, Y.~Ye, R.~Wang, C.~Liu, and K.~Roy, ``Going deeper in spiking neural networks: Vgg and residual architectures,'' \emph{Frontiers in neuroscience}, vol.~13, p.~95, 2019.

\bibitem{lu2020exploring}
S.~Lu and A.~Sengupta, ``Exploring the connection between binary and spiking neural networks,'' \emph{Frontiers in neuroscience}, vol.~14, 2020.

\bibitem{fgsm_goodfellow2014explaining}
I.~J. Goodfellow, J.~Shlens, and C.~Szegedy, ``Explaining and harnessing adversarial examples,'' \emph{arXiv preprint arXiv:1412.6572}, 2014.

\bibitem{fgsm2_sharmin2019comprehensive}
S.~Sharmin, P.~Panda, S.~S. Sarwar, C.~Lee, W.~Ponghiran, and K.~Roy, ``A comprehensive analysis on adversarial robustness of spiking neural networks,'' in \emph{2019 International Joint Conference on Neural Networks (IJCNN)}.\hskip 1em plus 0.5em minus 0.4em\relax IEEE, 2019, pp. 1--8.

\bibitem{bd1_chen2017targeted}
X.~Chen, C.~Liu, B.~Li, K.~Lu, and D.~Song, ``Targeted backdoor attacks on deep learning systems using data poisoning,'' \emph{arXiv preprint arXiv:1712.05526}, 2017.

\bibitem{bd2_adi2018turning}
Y.~Adi, C.~Baum, M.~Cisse, B.~Pinkas, and J.~Keshet, ``Turning your weakness into a strength: Watermarking deep neural networks by backdooring,'' in \emph{27th USENIX Security Symposium (USENIX Security 18)}, 2018, pp. 1615--1631.

\bibitem{bd3_gao2019strip}
Y.~Gao, C.~Xu, D.~Wang, S.~Chen, D.~C. Ranasinghe, and S.~Nepal, ``Strip: A defence against trojan attacks on deep neural networks,'' in \emph{Proceedings of the 35th Annual Computer Security Applications Conference}, 2019, pp. 113--125.

\bibitem{madry2017towards}
A.~Madry, A.~Makelov, L.~Schmidt, D.~Tsipras, and A.~Vladu, ``Towards deep learning models resistant to adversarial attacks,'' \emph{arXiv preprint arXiv:1706.06083}, 2017.

\bibitem{achille2018critical}
A.~Achille, M.~Rovere, and S.~Soatto, ``Critical learning periods in deep networks,'' in \emph{International Conference on Learning Representations}, 2018.

\bibitem{kirkpatrick2017overcoming}
J.~Kirkpatrick, R.~Pascanu, N.~Rabinowitz, J.~Veness, G.~Desjardins, A.~A. Rusu, K.~Milan, J.~Quan, T.~Ramalho, A.~Grabska-Barwinska \emph{et~al.}, ``Overcoming catastrophic forgetting in neural networks,'' \emph{Proceedings of the national academy of sciences}, vol. 114, no.~13, pp. 3521--3526, 2017.

\bibitem{covi2021adaptive}
E.~Covi, E.~Donati, X.~Liang, D.~Kappel, H.~Heidari, M.~Payvand, and W.~Wang, ``Adaptive extreme edge computing for wearable devices,'' \emph{Frontiers in Neuroscience}, vol.~15, p. 611300, 2021.

\bibitem{gao2023presynaptic}
T.~Gao, B.~Deng, J.~Wang, and G.~Yi, ``Presynaptic spike-driven plasticity based on eligibility trace for on-chip learning system,'' \emph{Frontiers in Neuroscience}, vol.~17, p. 1107089, 2023.

\bibitem{pehle2022brainscales}
C.~Pehle, S.~Billaudelle, B.~Cramer, J.~Kaiser, K.~Schreiber, Y.~Stradmann, J.~Weis, A.~Leibfried, E.~M{\"u}ller, and J.~Schemmel, ``The brainscales-2 accelerated neuromorphic system with hybrid plasticity,'' \emph{Frontiers in Neuroscience}, vol.~16, p. 795876, 2022.

\bibitem{frenkel2022reckon}
C.~Frenkel and G.~Indiveri, ``Reckon: A 28nm sub-mm2 task-agnostic spiking recurrent neural network processor enabling on-chip learning over second-long timescales,'' in \emph{2022 IEEE International Solid-State Circuits Conference (ISSCC)}, vol.~65.\hskip 1em plus 0.5em minus 0.4em\relax IEEE, 2022, pp. 1--3.

\bibitem{liu2018memory}
C.~Liu, G.~Bellec, B.~Vogginger, D.~Kappel, J.~Partzsch, F.~Neum{\"a}rker, S.~H{\"o}ppner, W.~Maass, S.~B. Furber, R.~Legenstein \emph{et~al.}, ``Memory-efficient deep learning on a spinnaker 2 prototype,'' \emph{Frontiers in neuroscience}, vol.~12, p. 840, 2018.

\bibitem{fang2021deep}
W.~Fang, Z.~Yu, Y.~Chen, T.~Huang, T.~Masquelier, and Y.~Tian, ``Deep residual learning in spiking neural networks,'' \emph{Advances in Neural Information Processing Systems}, vol.~34, pp. 21\,056--21\,069, 2021.

\bibitem{hu2021spiking}
Y.~Hu, H.~Tang, and G.~Pan, ``Spiking deep residual networks,'' \emph{IEEE Transactions on Neural Networks and Learning Systems}, vol.~34, no.~8, pp. 5200--5205, 2021.

\bibitem{zhou2022spikformer}
Z.~Zhou, Y.~Zhu, C.~He, Y.~Wang, S.~Yan, Y.~Tian, and L.~Yuan, ``Spikformer: When spiking neural network meets transformer,'' \emph{arXiv preprint arXiv:2209.15425}, 2022.

\bibitem{zhang2022spiking}
J.~Zhang, B.~Dong, H.~Zhang, J.~Ding, F.~Heide, B.~Yin, and X.~Yang, ``Spiking transformers for event-based single object tracking,'' in \emph{Proceedings of the IEEE/CVF conference on Computer Vision and Pattern Recognition}, 2022, pp. 8801--8810.

\bibitem{bal2024spikingbert}
M.~Bal and A.~Sengupta, ``Spikingbert: Distilling bert to train spiking language models using implicit differentiation,'' in \emph{Proceedings of the AAAI conference on artificial intelligence}, vol.~38, no.~10, 2024, pp. 10\,998--11\,006.

\end{thebibliography}


\newpage
\appendices
\section{SNN Accuracy in Convolutional Architectures }
\label{app:perf}
In this section, the DVS Gesture dataset is classified by FF and REC SNNs consisting of 3 convolutional layers with 64, 128, and 128 channels respectively (each with a kernel size of 7, a stride of 1, and a padding of 1). The observations (Table \ref{tab:dvs}) are consistent with those in Section \ref{sec:perf}.

\begin{table}[htbp]
\caption{SNN accuracy (\%) is benchmarked on the DVS Gesture dataset, with results averaged across five independent runs. The observations align with prior findings. \label{tab:dvs}}
\begin{center}
\begin{tabular}{|c|c|c|}
\hline
\textbf{Training}&\multicolumn{2}{|c|}{\textbf{DVS Gesture}}\\
\cline{2-3} 
\textbf{Methods} & \textit{FF} & \textit{REC} \\
\hline
BPTT    & $91.74$  & $93.04$ \\
\hline
DECOLLE & $90.15$ & $91.28 $ \\ 
\hline
\end{tabular}
\label{tab:benchconv}
\end{center}
\end{table}

\section{Hyperparameter Configuration}
\label{app:param}
Table \ref{tab:param} summarizes the hyper-parameters for optimal performance achieved by each method, where LR represents learning rate and BS represents batch size.
\begin{table}[htbp]
\caption{Hyper-parameters for optimal performance on N-MNIST, DVS Gesture, and TIMIT datasets.\label{tab:param}}
\begin{center}
\begin{tabular}{|l|c|c|c|c|c|c|}
\hline
\textbf{Model}& $e^{\frac{-1}{\tau_{\rm syn}}}$& $e^{\frac{-1}{\tau_{\rm mem}}}$& $\mathcal{V}_{\rm th}$ & LR & BS & Epochs\\
\hline
\multicolumn{7}{|c|}{\textbf{N-MNIST}} \\
    \hline
    BP FF & 0.9 & 0.5 & 0.9 & 1e-4 & 16 & 100\\
     e-prop FF & 0.99 & 0.95 & 0.2 & 5e-3 & 5 & 100\\
     DECOLLE FF & 0.97 & 0.92 & 1.0 & 1e-3 & 72 & 100\\
      BP REC & 0.9 & 0.5 & 0.9 & 1e-2 & 256 & 100\\
     e-prop REC & 0.99 & 0.95 & 0.2  & 5e-3 & 4 & 100\\
     DECOLLE REC & 0.97 & 0.92 & 1.0 & 1e-5 & 72 & 100\\
    \hline
\multicolumn{7}{|c|}{\textbf{DVS-Gesture}} \\
    \hline
    BP FF & 0.9 & 0.5 & 1.0 & 1e-3 & 16 & 100\\
     e-prop FF & 0.95 & 0.65 & 0.3 & 1e-3 & 15 & 100\\
     DECOLLE FF & 0.9 & 0.65 & 0.9 & 2e-4 & 72 & 100\\
    BP REC & 0.95 & 0.5 & 0.9 & 1e-3 & 32 & 100\\ 
     e-prop REC & 0.95 & 0.6 & 0.7 & 3e-3 & 15 & 100\\
     DECOLLE REC & 0.05 & 0.2 & 1.0 & 3e-5 & 72 & 100\\
    \hline
\multicolumn{7}{|c|}{\textbf{TIMIT}} \\
    \hline
    BP FF & 0.9 & 0.5 & 1.0 & 1e-3 & 8 & 100\\
     e-prop FF & 0.99 & 0.3 & 1.0 & 5e-4 & 4 & 100\\
     DECOLLE FF & 0.97 & 0.9 & 0.9 & 2e-4 & 128 & 100\\
    BP REC & 0.9 & 0.7 & 1.0 & 1e-3 & 8 & 100\\ 
     e-prop REC & 0.99 & 0.3 & 1.0 & 5e-4 & 4 & 100\\
     DECOLLE REC & 0.97 & 0.3 & 1.0 & 1e-5 & 32 & 100\\
    \hline
\end{tabular}
\end{center}
\end{table}

\end{document}